\documentclass[fleqn,10pt]{wlscirep}

\usepackage[utf8]{inputenc}
\usepackage[T1]{fontenc}
\usepackage{textcomp}
\usepackage{graphicx}
\usepackage{booktabs}
\usepackage{amsmath, amssymb}
\usepackage{siunitx}
\usepackage{bm}
\usepackage{url}
\usepackage{microtype}
\title{Ink Detection from Surface Topography of the Herculaneum Papyri}

\author[1,*]{Giorgio Angelotti}
\author[1,2]{Federica Nicolardi}
\author[1,3]{Paul Henderson}
\author[1,4]{W. Brent Seales}
\affil[1]{Vesuvius Challenge, USA}
\affil[2]{Università degli Studi di Napoli Federico II, Italy}
\affil[3]{University of Glasgow, Scotland, UK}
\affil[4]{EduceLab, University of Kentucky, USA}
\affil[*]{giorgio@scrollprize.org}

\begin{abstract}
Reading the Herculaneum papyri is challenging because both the scrolls and the ink, which is carbon-based, are carbonized. In X-ray radiography and tomography, ink detection typically relies on density- or composition-driven contrast, but carbon ink on carbonized papyrus provides little attenuation contrast. Building on the morphological hypothesis, we show that the \emph{surface morphology} of written regions contains enough signal to distinguish ink from papyrus. To this end, we train machine learning models on three-dimensional \emph{optical profilometry} from mechanically opened Herculaneum papyri to separate inked and uninked areas. We further quantify how lateral sampling governs learnability and how a native-resolution model behaves on coarsened inputs. We show that high-resolution topography alone contains a usable signal for ink detection. Diminishing segmentation performance with decreasing lateral resolution provides insight into the characteristic spatial scales that must be resolved on our dataset to exploit the morphological signal. These findings inform spatial resolution targets for morphology-based reading of closed scrolls through X-ray tomography.
\end{abstract}

\begin{document}
\flushbottom
\maketitle

\section*{Introduction}
The Herculaneum papyri are hundreds of carbonized books in the form of scrolls discovered in the second half of the eighteenth century in the ruins of the Roman city of Herculaneum, destroyed and buried by the eruption of Vesuvius in 79~AD. Having been carbonized and buried under volcanic ash for almost two millennia, the collection constitutes the only library preserved from Greco-Roman antiquity\cite{Sider2005Villa, LongoAuricchio2020Villa}. Despite several opening and unrolling methods applied to the papyri over almost three centuries, hundreds of volumes remain unread\cite{Seales2013cron, Nicolardi2024nonsvolti}. More than 800 inventoried items – entire scrolls and pieces of scrolls – are still rolled, and their extreme fragility makes noninvasive reading extraordinarily challenging \cite{mocella2015phase,tack2016lead}. Recent work has proposed virtual-unwrapping pipelines for X-ray computed tomography (CT) scans of sealed scrolls \cite{seales2016engedi,bukreeva2016virtual, nicolardi2024revealing, nicolardi2024phercparis4, henderson26spiral}. These methods geometrically flatten papyrus layers and, when paired with imaging that yields ink contrast, can extract text from closed scrolls \cite{seales2016engedi,bukreeva2016virtual}.
For carbon inks on carbonized papyrus, however, absorption contrast is intrinsically weak, motivating the use of advanced modalities such as phase-contrast tomography and X-ray fluorescence \cite{parker2019invisibility,brun2016metallic, bonnerot2020xrf}.
Reading the Herculaneum papyri is also challenging for scrolls that, despite their fragility, could be mechanically unwrapped. Because carbon inks are difficult to distinguish in the visible spectrum, complementary surface-sensitive methods, such as infrared imaging\cite{Booras1999multispectral, Tournie2019SWIR, Seales2023digital} and pulsed thermography, have recently revealed writing that is nearly invisible to the eye on Herculaneum fragments, supporting the premise that chemical surface properties carry a usable signal for carbon inks \cite{ceccarelli2025thermo}.
These techniques collectively show promise in enhancing weak, density-based ink signals to increase readability in both open trays and X-ray CT of closed scrolls.

In parallel, cultural-heritage studies increasingly exploit optical profilometry and microprofilometry to characterize surface topography with sub-micrometre lateral sampling and nanometric vertical repeatability \cite{mazzocato2024microprofilometry}. Profilometry is well established in paper science and forensic document analysis, where microrelief encodes writing processes and indented strokes \cite{borch2002profilometry,indentedProfilometry2020}. These observations suggest a complementary hypothesis for carbonized papyri, often termed the \emph{morphological hypothesis}\cite{stephen2023thesis}, that ink deposition alters local microrelief sufficiently that a purely \emph{morphological} signal can discriminate inked and uninked regions, independent of composition. Consistent with this idea, preliminary SEM observations on the Smithsonian scroll have suggested differences in surface roughness between ink and substrate, with inked regions appearing smoother\cite{parker2019invisibility}.

In the current work, we test this hypothesis on mechanically unrolled fragments of carbonized Herculaneum papyri. Using a three-dimensional optical profilometer (Sensofar S~lynx~2) in confocal mode, we acquire calibrated heightmaps over individual letters. We normalize the heightmaps, convert them to images, and train a deep-learning segmentation model to classify ink versus papyrus using manually drawn labels from aligned brightfield photographs. We address three questions: (i) does topography alone, at high lateral sampling, suffice to detect ink? (ii) how does apparent performance change when the same samples are synthetically downsampled to coarser pixel sizes, and when a high-resolution model is evaluated on upsampled, degraded inputs? and (iii) can the morphological signal learned on some inked papyri generalize to an unobserved sample?

\noindent More specifically, the contributions of this work are:

\begin{itemize}
\item \textbf{Dataset.} An annotated optical-profilometry dataset spanning 14 regions across PHerc.~248, PHerc.~250, and PHerc.~500P2, comprising 16 individual letters captured at \SI{0.34}{\micro\metre} lateral sampling with \SI{8}{\nano\metre} vertical sensitivity.
\item \textbf{Topography-only segmentation.} A demonstration that \emph{surface topography alone} supports accurate ink segmentation at native lateral sampling (\SI{0.34}{\micro\metre}), with held-out Dice scores of approximately 0.88–0.90.
\item \textbf{Resolution–performance trend.} Evidence of a monotonic degradation with coarser sampling on this dataset: median Dice decreases from $0.890\,[0.864,0.905]$ at \SI{0.34}{\micro\metre} to $0.467\,[0.364,0.524]$ at \SI{10.88}{\micro\metre} (Friedman $\chi^{2}=88.34$, $p=1.01\times10^{-15}$; Page’s $L=3848$, one-sided $p=1.0\times10^{-4}$). This trend reflects our dataset and is not claimed as a universal bound.
\item \textbf{Cross-resolution inference.} A model trained at \SI{0.34}{\micro\metre} degrades in performance on upsampled coarse inputs; Dice falls gently by $\leq\!\SI{2.04}{\micro\metre}$ pixel sizes, but sharply to near zero by $\geq\!\SI{3.40}{\micro\metre}$, underscoring that a model trained on fine features still captures some signal up to $\SI{2.04}{\micro\metre}$, but this signal is completely lost soon after.
\item \textbf{Isotropic-voxel emulation.} Evaluating matched-resolution models on vertically discretized heightmaps (\(\Delta z=\Delta x\)) yields Dice values comparable to the continuous-height setting, suggesting that vertical quantization is not the primary bottleneck on this dataset.
\item \textbf{Cross-papyrus generalization.} A leave-one-papyrus-out evaluation at native sampling (\SI{0.34}{\micro\metre}) quantifies transfer across manuscripts. Aggregating held-out samples across splits yields a mean Dice of 0.691, while per-papyrus performance is heterogeneous (Table~\ref{tab:lopo_native}).
\end{itemize}

% -----------------------------------------------------------
% Methods
\section*{Methods}
\textbf{Samples.} We analyzed fragments from three carbonized papyri—PHerc.~248, PHerc.~250 and PHerc.~500P2. PHerc. 248 (Philodemus, \textit{De pietate}) and PHerc. 250 (Philodemus, \textit{De rhetorica} I) are the extant sheets of two \textit{scorze}––outermost portions of scrolls removed through longitudinal cuts in the 1750s––which were scraped off layer by layer in the first half of the 19th century. PHerc. 500P2 is one of the loose pieces inventoried under PHerc. number 500, which are currently stored in \textit{cassetto }28. Sixteen letters from these fragments were selected based on visible legibility and minimal physical damage. Each letter region measured approximately \SI{1.5}{\milli\metre} on a side.

\begin{figure*}[t]
  \centering
 \includegraphics[width=\textwidth]{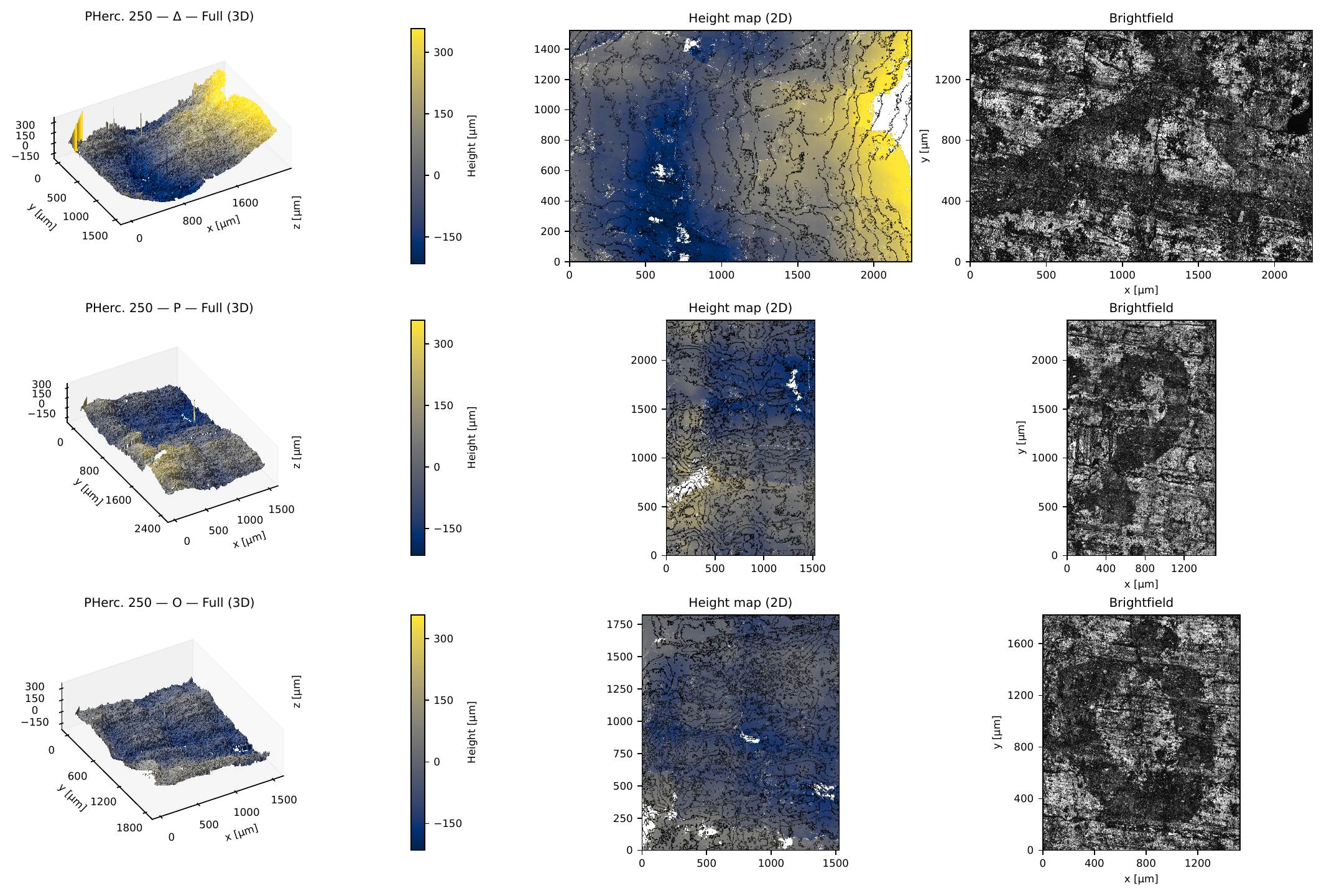}
  \caption{Representative profilometry overview for three samples from PHerc.~250. Each row shows a distinct letter. From left to right: (i) full 3D render of the measured topography; (ii) 2D heightmap with topographic contours; (iii) aligned brightfield image with visible ink. Axes are in \si{\micro\metre}; height color bars share the same scale within each row.}
  \label{fig:profilometry-overview}
\end{figure*}

\noindent\textbf{Acquisition.} Surface topography was measured using a Sensofar S~lynx~2 optical profilometer operating in confocal mode. The lateral sampling at native resolution was \SI{0.34}{\micro\metre}, and the vertical sensitivity was \SI{8}{\nano\metre}. For each letter, a co‑registered brightfield photograph was acquired to aid manual annotation. Examples of letters from PHerc.~250 are displayed in Figure~\ref{fig:profilometry-overview}.

\noindent\textbf{Preprocessing and synthetic downsampling.} Raw profilometry exports occasionally contained non-finite height values (NaN/Inf) at pixels where the confocal reconstruction failed. Before any resampling or normalization, we recorded a binary validity mask \(M(x)=\mathbb{1}[\mathrm{finite}(z(x))]\) and filled missing pixels to obtain dense maps compatible with nnU-Net training. Missing values were filled by inpainting \emph{only} the non-finite pixels using the fast-marching method of Telea\cite{telea2004inpainting} (OpenCV implementation; radius \(=3\) pixels). Inpainting was performed on a temporary 8-bit normalized map using robust linear scaling (0.5--99.5 percentiles) and then mapped back to the original height range. To guarantee that measured values were unchanged, all originally finite pixels were restored exactly after inpainting. heightmaps (native lateral sampling \SI{0.34}{\micro\metre}) were then normalized to the range [0,1] and converted from 64-bit floats to 16-bit unsigned integers. To quantify how segmentation performance depends on lateral sampling, and to emulate the coarser effective sampling that may be encountered in volumetric imaging of closed scrolls, we synthetically downsampled each heightmap. Coarser effective lateral resolutions were simulated by block averaging using square kernels of size $n\times n$ pixels where $n$ varied from 2 to 32. The downsampled maps were then upsampled back to the original grid using bilinear interpolation to create paired inputs for cross‑resolution inference. In a complementary “isotropic voxel” simulation, we additionally discretized the height values in physical units (\si{\micro\metre}) into uniform bins whose width matched the effective lateral pixel size at each resolution (\(\Delta z=\Delta x\)). These operations (downsampling, upsampling and z-binning) were applied \emph{before} per-sample normalization and uint16 conversion. We did not apply plane subtraction or tilt correction.

\noindent\textbf{Labelling and splits.} Binary ink masks were manually delineated on aligned brightfield photographs using GIMP. 16 letters are grouped into 14 independent samples. Most samples include a single letter. For all experiments, we used five-fold cross-validation over the 14 samples. In each fold, models were trained on approximately 80\% of samples and evaluated on the held-out 20\%, and results were aggregated such that each sample contributes exactly one held-out score per condition. For matched-resolution experiments, we trained separate 2D nnU-Net models at each lateral sampling. For cross-resolution inference, the model trained at \SI{0.34}{\micro\metre} was evaluated on degraded versions of the held-out inputs. For z-binning, the matched-resolution models were evaluated on z-binned versions of the held-out inputs, using the same fold-wise models trained on continuous heightmaps. To assess cross-papyrus generalization, we additionally performed a leave-one-papyrus-out evaluation at \SI{0.34}{\micro\metre}: for each papyrus in turn, we trained on all samples from the other two papyri and evaluated on all samples from the held-out papyrus.

\noindent\textbf{Model and training.} We used the nnU-Net framework \cite{isensee2021nnU-Net} with a U‑Net‑like encoder–decoder architecture and default hyperparameters. Training used the Dice plus cross‑entropy loss, the Adam optimizer, a learning rate of $10^{-3}$, and data augmentation consisting of random flips, rotations, intensity scaling, and elastic deformations. Models were trained with early stopping on convergence of the validation loss.

\noindent\textbf{Evaluation and statistics.} Segmentation performance was quantified using the Dice coefficient on held‑out samples. For matched‑resolution models, we reported Dice across the 14 held‑out samples aggregated over cross‑validation folds. Analyses were performed on related samples across pixel sizes. Because normality could not be assumed, we used non-parametric tests: Friedman test (omnibus, two-tailed) and Page’s L (ordered trend, one-tailed). Unless otherwise noted, we used a significance threshold $\alpha=0.05$. Post-hoc pairwise comparisons used the Wilcoxon signed-rank test (two-tailed) with Holm correction controlling the family-wise error rate across all pixel-size contrasts. We summarize data with median ([Q1, Q3]) and mean $\pm$ standard deviation. To assess whether missing-value patterns could confound learning, we also quantified the fraction of non-finite pixels in the raw (pre-fill) heightmaps inside ink and papyrus regions and compared paired fractions with a Wilcoxon signed-rank test; additionally, we report the Dice overlap between the missingness mask and ink annotations as a trivial mask-only baseline.

\section*{Results}

\subsection*{Overview}
In total, we analyzed 16 letters grouped in 14 samples across three papyri. The principal outcomes of the segmentation experiments are illustrated in Figures~\ref{fig:photo_grid} and~\ref{fig:dice_vs_pixel}, Table~\ref{tab:merged_by_pixel_size}, and the leave-one-papyrus-out generalization summary in Table~\ref{tab:lopo_native}. High‑resolution topography supported accurate ink segmentation, with median held‑out Dice scores around 0.89. Performance degraded monotonically when the lateral sampling was synthetically coarsened by block averaging, and a model trained on native sampling retained reasonable accuracy only up to approximately \SI{1}{\micro\metre} effective pixel size. When presented to the high‑resolution model, coarser inputs showed high performance up to \SI{1.36}{\micro\metre} lateral sampling but degraded at coarser resolution. For visual guidance, we show \(\mathrm{Dice}=0.70\) as a commonly used reference overlap threshold in segmentation\cite{zijdenbos1994morphometric}.
At native sampling, leave-one-papyrus-out evaluation yields a pooled mean Dice of 0.691 across all held-out samples, with substantial variation by papyrus (Table~\ref{tab:lopo_native}).

\subsection*{Qualitative examples}
\begin{figure}[htbp]
  \centering
  \includegraphics[width=0.75\linewidth]{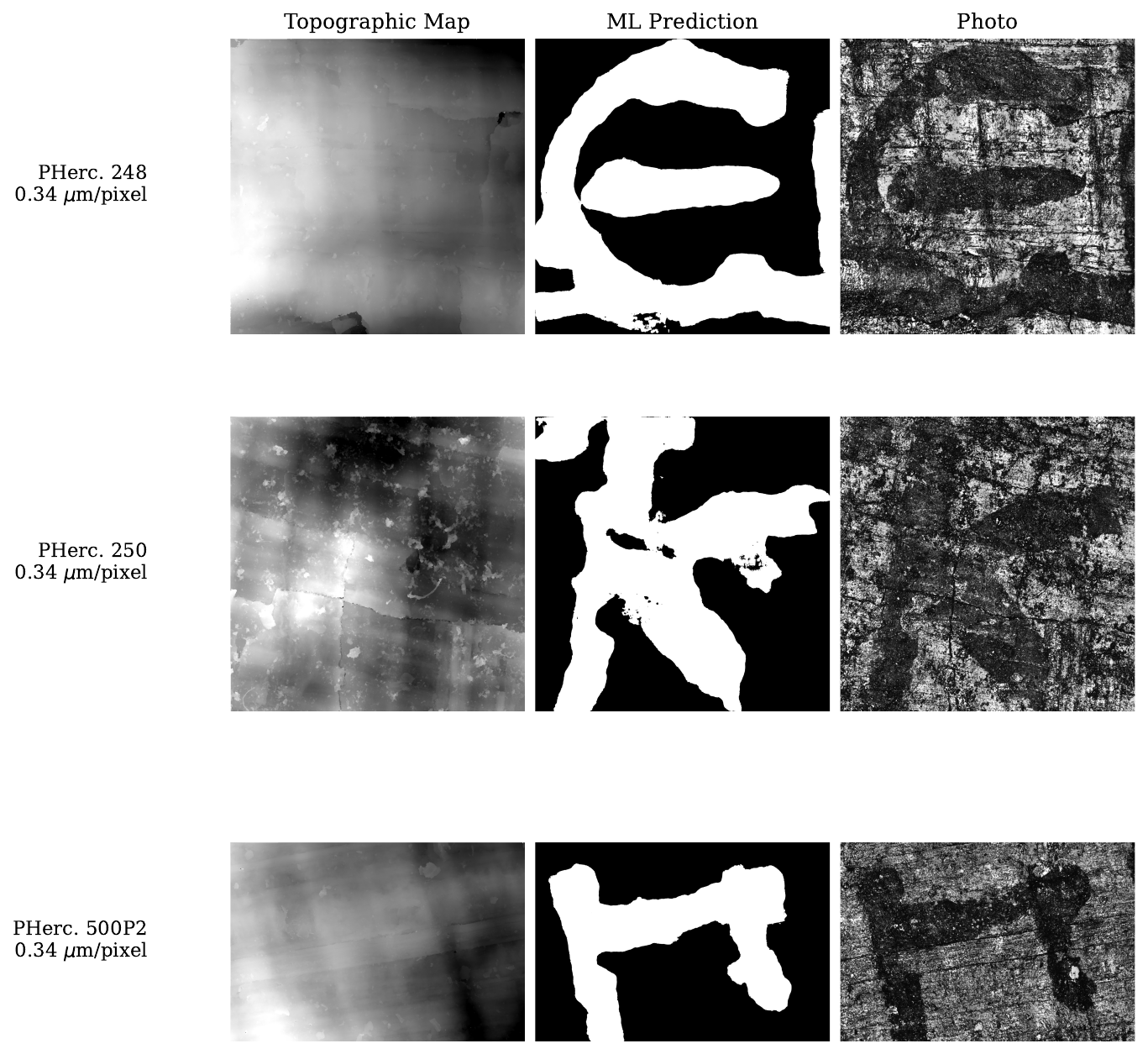}
  \caption{Qualitative examples across three papyri. Rows from top to bottom are $\epsilon$ on PHerc.~248, $\kappa$ on PHerc.~250, and $\eta$ on PHerc.~500P2. Columns from left to right are (i) topographic map converted to uint16 image,  (ii) nnU-Net model prediction, and (iii) brightfield photo. The grid structure visible in the heightmap is the lattice of fibers composing the papyrus substrate.}
  \label{fig:photo_grid}
\end{figure}

Figure~\ref{fig:photo_grid} illustrates three representative letters: $\epsilon$ from PHerc.~248, $\kappa$ from PHerc.~250, and $\eta$ from PHerc.~500P2. From left to right, columns show the topographic 16-bit unsigned-integer image after conversion, the predicted ink mask, and the co-registered brightfield photograph. The ink is not directly visible in the topographic image. In the full dataset, the heightmap appears to be dominated by the inclination and curvature of the surface, which is not perfectly flat, and by the shape of fibers that compose the papyrus sheet. Indeed, the surface can have some valleys or hills juxtaposed with papyrus fibers. Nevertheless, the trained machine learning model closely tracks stroke interiors and tapers at feathered boundaries where height relief approaches the background micro-roughness.

\noindent\textbf{Learning from topography alone.}
At the native sampling of \SI{0.34}{\micro\metre}, training converged smoothly across folds. Held-out Dice scores clustered around 0.88–0.90 with a median of $0.890\,[0.864,0.905]$, supporting the central claim that surface morphology alone provides a discriminative signal for ink and validating the \emph{morphological hypothesis}.

\noindent\textbf{Missing-data artifact controls.}
Across the \(n=14\) samples at native sampling, the raw (pre-fill) heightmaps contained a small fraction of non-finite pixels (median \(1.665\%\,[1.250\%,2.352\%]\)). Missingness was not systematically elevated inside ink relative to papyrus (ink: \(1.653\%\,[1.443\%,2.357\%]\); papyrus: \(1.768\%\,[1.075\%,2.379\%]\); paired Wilcoxon signed-rank test: \(W=52\), two-sided \(p=1.0\)). The direct overlap between the missingness mask and ink annotations was negligible (Dice \(=0.032\,[0.028,0.044]\)), suggesting that non-finite-pixel dropout patterns do not trivially coincide with ink regions on this dataset.

\begin{figure}[htbp]
  \centering
  \includegraphics[width=\linewidth]{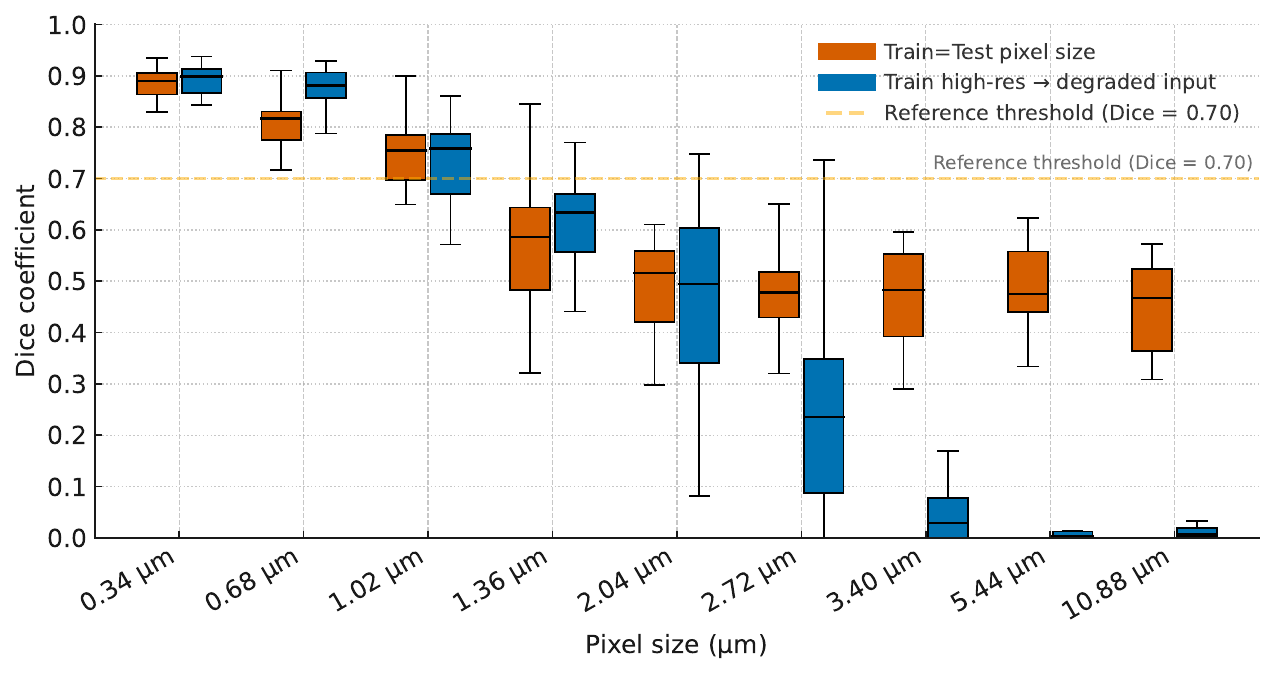}
    \caption{Segmentation performance (Dice) as a function of effective lateral sampling (pixel size, \si{\micro\metre}). Orange boxes: nnU-Net models trained and evaluated at matched resolution. Blue boxes: a single nnU-Net trained at \SI{0.34}{\micro\metre} evaluated on inputs downsampled to coarser pixel sizes and upsampled to the native grid. Boxes show the median and interquartile range across $n=14$ cases; whiskers extend to the most extreme non-outlier values (Tukey, $1.5\times\mathrm{IQR}$; outliers omitted). The dashed horizontal line marks \(\mathrm{Dice}=0.70\) as a reference overlap threshold.}
  \label{fig:dice_vs_pixel}
\end{figure}

\noindent\textbf{Resolution sensitivity with matched training and testing.}
The orange boxes in Figure~\ref{fig:dice_vs_pixel} and the first column partition of Table~\ref{tab:merged_by_pixel_size} summarize performance as lateral sampling is coarsened. Dice tended to decrease as pixel size increased, despite minor non-monotonic differences between resolutions. The Friedman test was highly significant ($\chi^{2}\!=\!88.34$, $p\!=\!1.01\times10^{-15}$), and Page’s $L$ confirmed an ordered decline ($L\!=\!3848$, one-sided $p\!=\!1.0\times10^{-4}$). Wilcoxon signed-rank pairwise comparisons involving \SI{0.34}{\micro\metre} remained significant after Holm adjustment (adjusted $p\le 0.0044$), with median paired changes in Dice ranging from $-0.086$ to $-0.417$ relative to \SI{0.34}{\micro\metre}. %The comparison between \SI{0.68}{\micro\metre} and \SI{1.02}{\micro\metre} was also significant (adjusted $p\!=\!0.0183$).
Pixel sizes at and above roughly \SI{2}{\micro\metre} were mutually indistinguishable after correction (adjusted $p\ge 0.245$). We interpret these as dataset-level observations rather than universal thresholds, as apparent bounds could shift with dataset size, variety, and machine learning model architecture.

\begin{table*}[htbp]
\centering
\caption{Dice versus lateral pixel size across three settings: (i) pixel size sensitivity (nnU-Net trained and evaluated at matched resolutions); (ii) a model trained at \SI{0.34}{\micro\metre} evaluated on degraded input; and (iii) matched-resolution evaluation on z-binned heightmaps (\(\Delta z=\Delta x\)) to emulate isotropic voxels. Values are median [Q1, Q3] and mean $\pm$ s.d. Only models at or finer than \SI{1.02}{\micro\metre} exceed the \(\mathrm{Dice}=0.70\) reference threshold.}
\label{tab:merged_by_pixel_size}
\small
\setlength{\tabcolsep}{4pt}
\begin{tabular}{@{}lcccccc@{}}
\toprule
\textbf{Lateral pixel size} &
\multicolumn{2}{c}{\textbf{Pixel size sensitivity (n=14)}} &
\multicolumn{2}{c}{\textbf{Model trained at \SI{0.34}{\micro\metre}}} &
\multicolumn{2}{c}{\textbf{Z-binned test inputs ($\Delta z=\Delta x$, n=14)}} \\
\cmidrule(lr){2-3}\cmidrule(lr){4-5}\cmidrule(l){6-7}
& \textbf{Median [Q1, Q3]} & \textbf{Mean $\pm$ s.d.}
& \textbf{Median [Q1, Q3]} & \textbf{Mean $\pm$ s.d.}
& \textbf{Median [Q1, Q3]} & \textbf{Mean $\pm$ s.d.} \\
\midrule
\SI{0.34}{\micro\metre}  & 0.890 [0.864, 0.905] & 0.881 $\pm$ 0.039 & 0.899 [0.866, 0.913] & 0.889 $\pm$ 0.038 & 0.864 [0.808, 0.900] & 0.839 $\pm$ 0.075 \\
\SI{0.68}{\micro\metre}  & 0.817 [0.775, 0.831] & 0.803 $\pm$ 0.068 & 0.881 [0.857, 0.906] & 0.877 $\pm$ 0.042 & 0.817 [0.755, 0.843] & 0.792 $\pm$ 0.085 \\
\SI{1.02}{\micro\metre}  & 0.754 [0.698, 0.784] & 0.748 $\pm$ 0.068 & 0.758 [0.670, 0.786] & 0.736 $\pm$ 0.091 & 0.742 [0.701, 0.787] & 0.744 $\pm$ 0.080 \\
\midrule
\SI{1.36}{\micro\metre}  & 0.586 [0.483, 0.644] & 0.570 $\pm$ 0.143 & 0.634 [0.557, 0.670] & 0.611 $\pm$ 0.104 & 0.596 [0.507, 0.645] & 0.586 $\pm$ 0.139 \\
\SI{2.04}{\micro\metre}  & 0.516 [0.421, 0.559] & 0.493 $\pm$ 0.093 & 0.494 [0.341, 0.604] & 0.455 $\pm$ 0.208 & 0.499 [0.451, 0.541] & 0.486 $\pm$ 0.083 \\
\SI{2.72}{\micro\metre}  & 0.478 [0.429, 0.518] & 0.480 $\pm$ 0.096 & 0.235 [0.088, 0.348] & 0.264 $\pm$ 0.234 & 0.486 [0.426, 0.515] & 0.473 $\pm$ 0.085 \\
\SI{3.40}{\micro\metre}  & 0.483 [0.392, 0.552] & 0.460 $\pm$ 0.106 & 0.029 [0.000, 0.078] & 0.053 $\pm$ 0.063 & 0.480 [0.381, 0.516] & 0.451 $\pm$ 0.106 \\
\SI{5.44}{\micro\metre}  & 0.475 [0.440, 0.558] & 0.482 $\pm$ 0.091 & 0.003 [0.000, 0.013] & 0.013 $\pm$ 0.019 & 0.474 [0.415, 0.534] & 0.475 $\pm$ 0.084 \\
\SI{10.88}{\micro\metre} & 0.467 [0.364, 0.524] & 0.450 $\pm$ 0.090 & 0.007 [0.003, 0.019] & 0.018 $\pm$ 0.024 & 0.461 [0.397, 0.534] & 0.452 $\pm$ 0.091 \\
\bottomrule
\end{tabular}
\end{table*}

\noindent\textbf{Isotropic-voxel emulation via z-binning.}
To approximate volumetric scans with isotropic voxels, we discretized height values into bins of width equal to the effective lateral pixel size at each resolution (\(\Delta z=\Delta x\)). The rightmost column group in Table~\ref{tab:merged_by_pixel_size} summarizes performance when the matched-resolution models are evaluated on z-binned held-out test inputs. Across resolutions, z-binning produced Dice values close to the continuous-height setting (leftmost column group), with absolute changes in median Dice of at most $0.026$. A similar decline with increasing pixel size persists.

\noindent\textbf{Cross-resolution generalization with a fixed high-resolution model.}
We trained a 2D nnU-Net model at \SI{0.34}{\micro\metre} and, at test time, degraded inputs by block-averaging to coarser effective resolutions before up-sampling back to the native grid. The blue boxes in Figure~\ref{fig:dice_vs_pixel} and the second column partition of Table~\ref{tab:merged_by_pixel_size} summarize the outcome. Performance remained high through \(\sim\)\SI{1}{\micro\metre}, declined by \SI{2.04}{\micro\metre}, and collapsed to near zero by \SI{3.40}{\micro\metre} and coarser. The drop is steeper than in the matched-train/matched-test setting, consistent with the loss of high-frequency morphological content and distribution shift introduced by down-sample–up-sample pipelines. Statistics were $\chi^{2}\!=\!104.61$, $p\!=\!4.86\times10^{-19}$ (Friedman) and $L\!=\!3947$, one-sided $p\!=\!1.0\times10^{-4}$ (Page’s trend test).

\noindent\textbf{Cross-papyrus generalization.}
To probe whether learned morphological cues transfer across manuscripts, we performed a leave-one-papyrus-out evaluation at the native sampling (\SI{0.34}{\micro\metre}). For each papyrus in turn, an nnU-Net was trained on all samples from the other two papyri and evaluated on the held-out papyrus. Aggregating all held-out samples across the three splits (\(n=14\)), the mean Dice was 0.691. However, per-papyrus results varied substantially: performance remained relatively high when holding out PHerc.~248 (median Dice 0.757) or PHerc.~250 (median Dice 0.817), but dropped markedly when holding out PHerc.~500P2 (median Dice 0.475), indicating heterogeneous cross-manuscript transfer on this dataset (Table~\ref{tab:lopo_native}).

\begin{table}[htbp]
\centering
\caption{Leave-one-papyrus-out generalization at native lateral sampling (\SI{0.34}{\micro\metre}). For each held-out papyrus, an nnU-Net was trained on all samples from the other two papyri and evaluated on all samples from the held-out papyrus. Values are median [Q1, Q3] and mean $\pm$ s.d. across samples from the held-out papyrus. Pooled across all held-out samples (\(n=14\)), the mean Dice is 0.691.}
\label{tab:lopo_native}
\small
\setlength{\tabcolsep}{6pt}
\begin{tabular}{@{}lccc@{}}
\toprule
\textbf{Held-out papyrus} & \textbf{n} & \textbf{Median [Q1, Q3]} & \textbf{Mean $\pm$ s.d.} \\
\midrule
PHerc.~248    & 5 & 0.757 [0.733, 0.812] & 0.777 $\pm$ 0.059 \\
PHerc.~250    & 5 & 0.817 [0.755, 0.822] & 0.789 $\pm$ 0.053 \\
PHerc.~500P2  & 4 & 0.475 [0.454, 0.482] & 0.461 $\pm$ 0.035 \\
\bottomrule
\end{tabular}
\end{table}
% -----------------------------------------------------------
% Discussion
\section*{Discussion}
Our results establish that high–resolution surface morphology carries sufficient signal to accurately segment carbon inks on carbonized papyri. Moreover, on our setup, performance decreases systematically when lateral sampling is degraded. A notable observation in Table~\ref{tab:merged_by_pixel_size} is that a model trained at the native \SI{0.34}{\micro\metre} sampling and evaluated on inputs blurred and up‑sampled to \SI{0.68}{\micro\metre} achieved higher Dice than models trained natively at the matched coarse resolution. This counterintuitive behavior echoes reports from the computer‑vision literature: when images are uniformly downsampled at the input or early in the network, small features can disappear before the model learns to detect them\cite{marin2019efficient,bencevic2023segment}. Training on high‑resolution data allows the network to learn morphological cues at full scale, which can still be exploited after moderate blurring. In contrast, training natively on coarse inputs forces the encoder to consume already low‑pass signals, leading to poorer segmentation. However, once effective pixel sizes exceed $\sim$\SI{1}{\micro\metre}, both strategies collapse, consistent with small‑object detection theories that emphasize information loss from downsampling\cite{li2024mban}. These considerations emphasize the need to preserve fine structure throughout the pipeline and caution against aggressive downsampling. The systematic decline in segmentation performance with increasing pixel size suggests that the discriminative morphological signal is concentrated at fine spatial scales and is progressively lost as resolution coarsens. In topographic maps, these fine-scale oscillations correspond to surface roughness; yet filtering to isolate roughness alone does not reliably separate inked letters from the substrate. This suggests that the convolutional models leverage a composite signal: surface roughness together with pressure-induced deformations from ink deposition, encoded in shape patterns that classical profile-analysis algorithms do not readily capture.

We also performed complementary experiments where we discretized the vertical height values into bins as large as the lateral pixels to emulate an isotropic voxel sampling, akin to what might be achievable with volumetric X‑ray tomography. The z-binned columns in Table~\ref{tab:merged_by_pixel_size} show that this z-quantization changes Dice only modestly across our dataset (maximum median shift $0.026$ across resolutions) and preserves the overall trend of declining performance with coarser sampling. These observations suggest that volumetric scans delivering effective isotropic resolution at or below \SI{1}{\micro\metre} could be sufficient for morphology‑based ink detection.

The study is affected by the following limitations. Our dataset is small (sixteen letters across fourteen samples from three papyri) and therefore statistical power is limited, particularly at the coarsest samplings where variance increases. Although we report leave-one-papyrus-out generalization at \SI{0.34}{\micro\metre} (Table~\ref{tab:lopo_native}), each papyrus contributes only 4–5 samples; thus, cross-papyrus estimates are preliminary and likely sensitive to sample selection and preservation state. Ground‑truth masks were derived from co‑registered brightfield photographs; this can introduce bias at feathered ink edges and propagate registration errors into evaluation. Synthetic downsampling by block averaging approximates but does not replicate the point‑spread and modulation‑transfer functions of real instruments, so our trends should be interpreted as qualitative rather than universal. A single encoder–decoder architecture and default training regimen were used; alternative losses, receptive‑field sizes, or multi‑scale designs may affect performance at coarse sampling. These limitations notwithstanding, the observed monotonic decline with increased pixel size and the steep degradation when a high‑resolution model is evaluated on substantially degraded inputs provide important guidance for the design of future experiments.
Finally, we assessed whether confocal reconstruction dropouts could explain segmentation performance. Non-finite pixels were rare (\(\sim\)1--2\%) and showed no systematic association with ink, and the missingness mask had negligible overlap with ink annotations (Dice \(\approx0.03\)). This indicates that simple missing-value patterns are unlikely to account for the observed topography-based segmentation; however, systematic height reconstruction biases correlated with surface optics remain a potential confound to be tested in future work.

\section*{Conclusions and Perspectives}
Optical profilometry with sub-micrometre lateral sampling enables accurate segmentation of carbon inks on carbonized papyri using surface morphology alone. On our dataset, accuracy declined as lateral sampling was coarsened: models trained and tested at matched resolutions lost performance, and a single high-resolution model evaluated on upsampled coarse inputs degraded even more steeply. On our dataset, reliable segmentation benefits from lateral sampling on the order of \SI{1}{\micro\metre} or finer. The exact requirement will depend on data quantity and diversity, signal-to-noise, model architecture, and critically on the amount and preservation state of ink (e.g., variation in deposited material across hands and possible effacement on long-exposed opened fragments).

If volumetric imaging of closed scrolls can deliver comparable \emph{effective} resolution at the layer surfaces, preserving similar microrelief and signal-to-noise ratio, then morphology-only detection becomes plausible inside sealed scrolls. We emphasize “effective” rather than nominal pixel size because point-spread, partial-volume effects, scattering, reconstruction regularization, and other artifacts can all diminish resolvable detail. Despite these caveats, the present results refine and extend morphology-assisted reading efforts by quantifying how effective surface resolution controls topography-only ink detectability. They motivate future imaging campaigns that preserve papyrus microrelief and target higher effective resolution in volumetric scans of sealed scrolls, where morphology-based cues remain detectable.

\section*{Data Availability}
Data samples, labels, fold splits, and configuration files will be released in a public repository upon publication. Until then, they are available from the corresponding author for peer review on request.

% =========================================================
% Bibliography (self-contained)

\section*{Acknowledgements}
We thank the curators and conservators responsible for the PHerc.~material at the National Library of Naples "Vittorio Emanuele III" for access and guidance, in particular Silvia Scipioni and Giovanni Bova. This research was supported by the Vesuvius Challenge (https://scrollprize.org), a non-profit donation-funded organization that aims to read the full collection of sealed scrolls from Herculaneum. We are particularly grateful to Nat Friedman for initiating and supporting the Vesuvius Challenge. We are also grateful to the previous work by the EduceLab, University of Kentucky, sponsored by the Mellon Foundation, that laid the groundwork for the realization of this research.

\section*{Author contributions statement}
G.A. conceived the study, collected the data, performed the analyses, and drafted the manuscript. F.N. selected and identified the papyri and provided papyrological expertise. P.H. provided computer vision and machine learning input. W.B.S. provided input on the state-of-the-art, related work and presentation of results. All authors reviewed and approved the manuscript.

\section*{Additional information}
This research did not receive a formal research grant from any funding agency. Support was provided through the Vesuvius Challenge, a donation-funded, non-profit initiative. The authors declare no competing interests.

\end{document}